\documentclass[runningheads]{llncs}
\usepackage{graphicx}
\usepackage{tikz}
\usepackage{comment}
\usepackage{amsmath,amssymb} % define this before the line numbering.
\usepackage{color}
\usepackage{orcidlink}

\usepackage{booktabs}
\usepackage{multirow}
\usepackage[normalem]{ulem}
\useunder{\uline}{\ul}{}
\usepackage{bbding}
\usepackage{subfig}
\usepackage[accsupp]{axessibility}  % Improves PDF readability for those with disabilities.

\begin{document}
% \linenumbers
\pagestyle{headings}
\mainmatter
\def\ECCVSubNumber{2488}  % Insert your submission number here

\title{Can Shuffling Video Benefit Temporal Bias Problem: A Novel Training Framework for Temporal Grounding} % Replace with your title

% \begin{comment}
% \titlerunning{ECCV-22 submission ID \ECCVSubNumber} 
% \authorrunning{ECCV-22 submission ID \ECCVSubNumber} 
% \author{Anonymous ECCV submission}
% \institute{Paper ID \ECCVSubNumber}
% \end{comment}
%******************

%\begin{comment}
\titlerunning{Can Shuffling Video Benefit Temporal Bias Problem}
% \newcounter{savecntr}% Save footnote counter
% \newcounter{restorecntr}% Restore footnote counter
\author{Jiachang Hao \and
Haifeng Sun$^\star$ \and
Pengfei Ren \and
Jingyu Wang\thanks{Corresponding author} \and
Qi Qi \and
Jianxin Liao
}

\authorrunning{J. Hao et al.}
\institute{State Key Laboratory of Networking and Switching Technology,\\
Beijing University of Posts and Telecommunications\\
\email{\{haojc, hfsun, rpf, wangjingyu, qiqi8266\}@bupt.edu.cn}
\email{jxlbupt@gmail.com}
}
%\end{comment}
%******************

\maketitle

\begin{abstract}
Temporal grounding aims to locate a target video moment that semantically corresponds to the given sentence query in an untrimmed video. However, recent works find that existing methods suffer a severe temporal bias problem. These methods do not reason the target moment locations based on the visual-textual semantic alignment but over-rely on the temporal biases of queries in training sets. To this end, this paper proposes a novel training framework for grounding models to use shuffled videos to address temporal bias problem without losing grounding accuracy. Our framework introduces two auxiliary tasks, cross-modal matching and temporal order discrimination, to promote the grounding model training. The cross-modal matching task leverages the content consistency between shuffled and original videos to force the grounding model to mine visual contents to semantically match queries. The temporal order discrimination task leverages the difference in temporal order to strengthen the understanding of long-term temporal contexts. Extensive experiments on Charades-STA and ActivityNet Captions demonstrate the effectiveness of our method for mitigating the reliance on temporal biases and strengthening the model's generalization ability against the different temporal distributions. Code is available at \url{https://github.com/haojc/ShufflingVideosForTSG}.
\end{abstract}

\section{Introduction}
\label{sec:intro}
\begin{figure}[!t]
 \centering
 \begin{subfloat}[\textbf{Temporal grounding task}]{\includegraphics[width=0.65\textwidth]{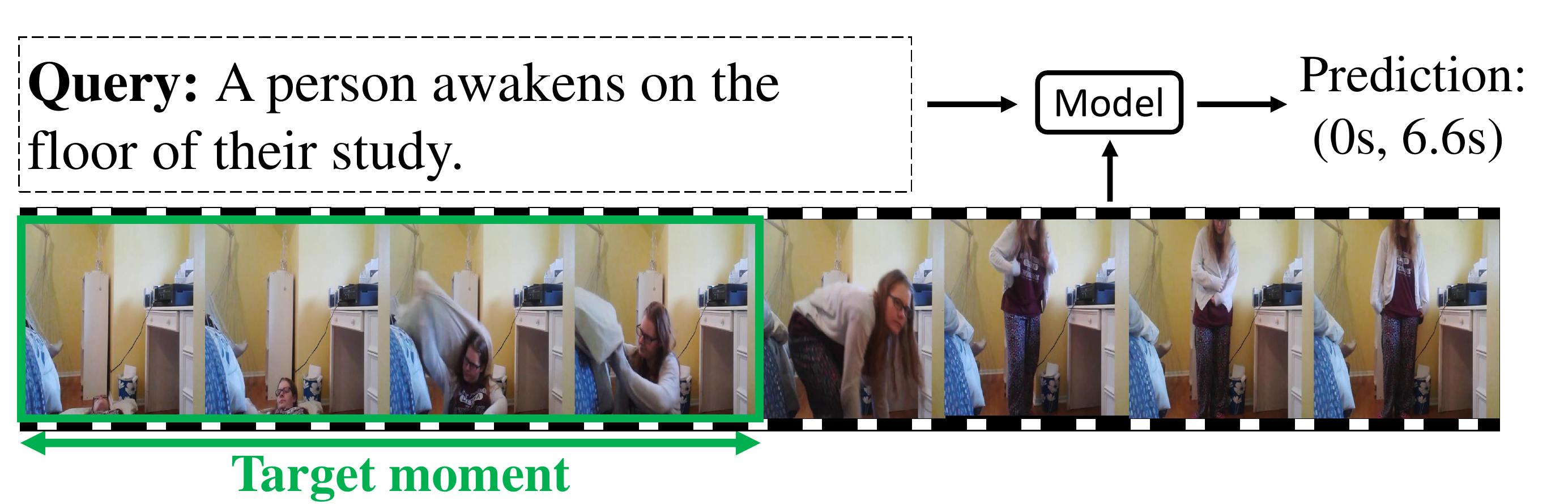}%
  \label{fig:TSG_example}}
 \end{subfloat}
 \begin{subfloat}[\textbf{Temporal bias problem}]{\includegraphics[width=0.65\textwidth]{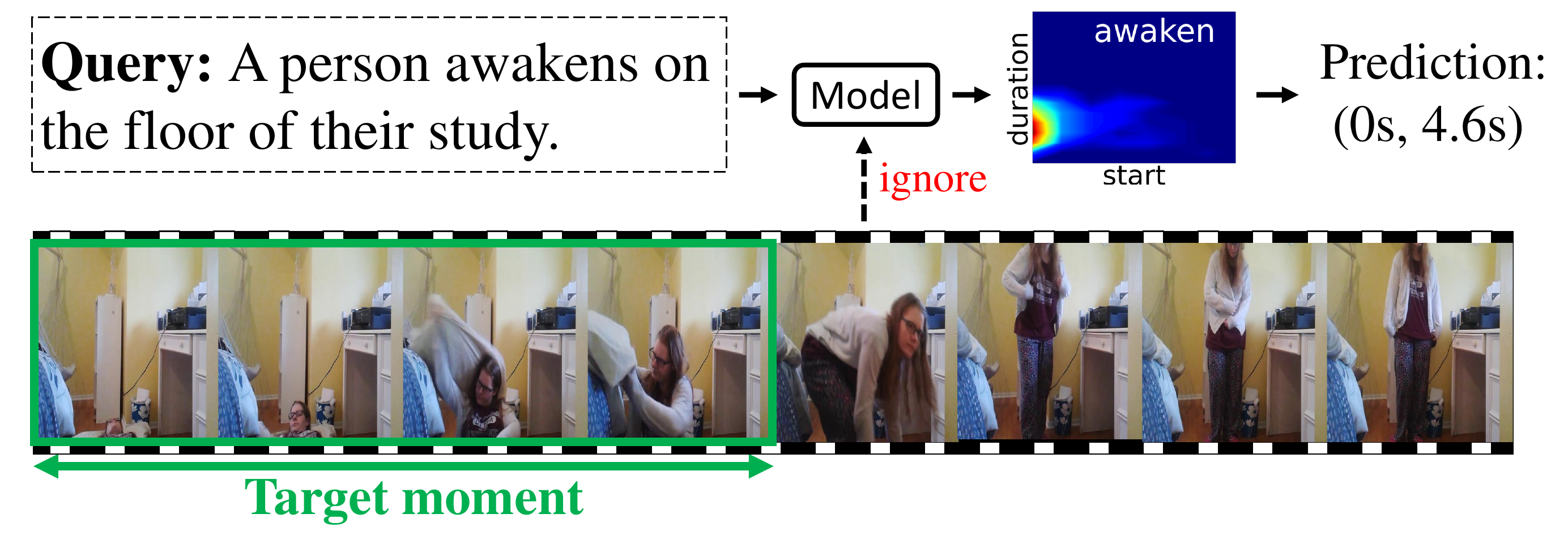}%
  \label{fig:TBP}}
 \end{subfloat}
 \caption{\textbf{(a)} Temporal grounding is to localize a moment with the start point (0s) and end point (6.6s) in the video for the query. \textbf{(b)} An example of temporal bias problem: a model ignores the visual input and uses the memorized temporal bias of the word `awaken' in the training set of Charades-STA to make the prediction.}
 \label{fig:TSG_SG_example}
 \end{figure}

Temporal grounding~\cite{tall2017,mcn2017} aims to localize the relevant video moment of interest semantically corresponding to the given sentence query in an untrimmed video, as illustrated in Fig.~\ref{fig:TSG_example}. Due to its vast potential applications in video captioning, video question answering, and video retrieval, this task has attracted increasing interest over the last few years. However, this task suffers a temporal bias problem~\cite{bmvc2020,closelook2020,dcm2021}, which severely hinders the development of the temporal grounding task.

Temporal bias problem refers to that a method reasons the target moment locations not based on the visual-textual semantic matching but over-relies on the temporal biases of queries in training set~\cite{bmvc2020}. As illustrated in Fig.~\ref{fig:TBP}, a grounding model ignores visual inputs and takes a shortcut to directly exploit the memorized temporal biases of the given query in training set to reason the location of the target video moment. The temporal bias problem severely hinders the development of temporal grounding~\cite{bmvc2020,dcm2021}. Because when we give a sentence query, we wish the target moment to be localized based on the query semantics wherever the target moment is. \cite{bmvc2020,closelook2020} found that many state-of-the-art methods~\cite{tall2017,ablr2019,2dtan2020,scdm2019,lg2020,vslnet2020,drn2020} suffer this problem and perform poor generalization ability against the different temporal distributions. Some methods~\cite{scdm2019,2dtan2020} even do not make any use of the visual input during reasoning. To this end, this paper aims to mitigate the excessive reliance on temporal biases and strengthen the model's generalization ability against the different temporal distributions.

Video-query pairs in existing datasets have high correlations between queries and ground-truth temporal positions of target moments, making it possible for a grounding model to take a shortcut~\cite{bmvc2020,dcm2021}. So we ask \emph{whether we can shuffle videos to break these correlations to address the temporal bias problem.} We can shift randomly target moments to other temporal positions in videos to dilute the temporal bias of the corresponding query in training set. Thus the shortcut of memorizing temporal biases will be ineffective, and the model has to turn the attention back to the visual contents semantically matching queries. However, directly using these shuffled videos as training samples is not appropriate. On the one hand, these shuffled videos may not match the real situation, causing a poor generalization ability of the grounding model. On the other hand, shuffling videos disturbs the long-term temporal contexts within videos. These incorrect contexts may weaken the perception ability of grounding models for long-term temporal relations, which is important for temporal grounding~\cite{2dtan2020,mllc2018}.

To this end, instead of using the shuffled videos as augmented training samples, we propose to take the shuffled and original videos as paired input and design auxiliary tasks to promote the grounding model training. We design two auxiliary tasks, cross-modal matching and temporal order discrimination, to mine the cross-modal semantic relevance from the paired videos and mitigate the reliance on temporal biases. The cross-modal matching task requires that the model predicts as consistent frame-level cross-modal relevance as possible for target moments, even if their temporal positions change. This task encourages the grounding model to focus on spatial and short-term visual contents to semantically match queries. The temporal order discrimination task is to discriminate whether the video moment sequence is in correct temporal order. This task guides the grounding model to learn the correct temporal contexts and thus strengthens the perception ability on long-term temporal relations.

We propose a span-based framework to handle the temporal grounding and two auxiliary tasks. And our method can be easily transferred to other grounding models. 
To sum up, the main contributions of our work are as follows:

(1) We propose a novel training framework for temporal grounding models to use shuffled videos to address the temporal bias problem.

(2) Our cross-modal matching task can mitigate the model's reliance on temporal biases and turn the attention back to the visual-textual semantic matching, and the temporal order discrimination task can strengthen the understanding of long-term temporal relations.

(3) Extensive experiments on Charades-STA and ActivityNet Captions demonstrate the effectiveness of our method for mitigating the grounding model's reliance on temporal biases and strengthening the generalization ability against the different temporal distributions. And we achieve state-of-the-art on the re-divided splits for temporal bias problem.

\section{Related Work}
\paragraph{Temporal Grounding}
Temporal grounding, also known as temporal sentence grounding in video and video moment retrieval, was first proposed by \cite{tall2017,mcn2017}. Proposal-based methods formulate this task as a ranking task to find the best matching video proposal for a given sentence query. These methods first generate the candidate video segments by slide windows \cite{tall2017,role2018,acrn2018} or a proposal network \cite{qspn2019,sap2019,bpnet2021,apgn2021} or predefined anchors \cite{tgn2018,tmn2018,man2019,cmin2019,cmin22020,csmgan2020,cbln2021,ianet2021,gtr2021,emai2022,mgsl2022}, and then semantically match each candidate with the sentence query. However, the proposal generation and semantic matching for all the proposals are resource-consuming and inefficient. To discard proposals, proposal-free methods encode the video modality only once and directly interact each video frame with the sentence query. 
% Proposal-free methods can further be divided into two groups, i.e., regression-based methods and span-based methods. 
Specifically, regression-based methods \cite{ablr2019,debug2019,pmiloc2020,hvtg2020,drft2021} regress the temporal coordinates of the localized video moment from a compact representation. Span-based methods \cite{excl2019,tmlga2020,gdp2020,abin2020,conquer2021,ivg2021,vslnet2,qave} predict the probabilities of each frame being the start/end of the location.

\paragraph{Temporal Bias Problem}
\cite{bmvc2020} first proposed the temporal bias problem. They found that the popular datasets for temporal grounding task include significant biases
through explicit statistics about temporal locations of the top-50 verbs in datasets. 
Then they 
% designed a series of experiments and 
verified that some state-of-the-art models did not achieve cross-modal alignment but exploited dataset biases instead.
% To correctly evaluate models' performance and generalization ability, 
To correctly evaluate grounding performance,
\cite{closelook2020} re-divided the splits of two popular datasets, Charades-STA and ActivityNet Captions, to make the training and test sets have different temporal distributions of queries.
To address the temporal bias problem, DCM~\cite{dcm2021} disentangles temporal position information from each proposal feature via a constraint loss and then leverages causal intervention 
% to force the model 
to fairly consider all candidate proposals. To evaluate the effectiveness, DCM~\cite{dcm2021} simulated the out-of-distribution test samples by inserting a sequence of generated video features at the beginning of the original video feature sequence. However, this simulation is not much convincing. On the one hand, the inserted features are generated from a normal distribution so that these features may not contain any meaningful content. On the other hand, the length of the inserted video sequence is too short to change the temporal biases significantly. In this paper, we use the re-divided splits~\cite{closelook2020} to evaluate performance and generalization ability.

\paragraph{Temporal Relation Modeling}
Temporal relation modeling is a fundamental problem in video understanding. Typical methods apply RNN~\cite{bss2015}, 3D-CNN~\cite{icml2010,kinetics,r21d} to capture the short-term temporal relations and non-local~\cite{nonlocal}, Transformer~\cite{vaswani2017attention}, TDN~\cite{tdn2021} to capture the long-term ones. However, due to the black box of neural networks, it is uncertain whether the features learned from the aforementioned mechanism contain the temporal relations or other information like scene, appearance, and temporal bias. Therefore, some works~\cite{o3d,shuffleandlearn,shuffleandatten,sortsequence2017,cliporder2019} attempt to explicitly strengthen video representation learning in temporal relations by using auxiliary tasks. Temporal order verification is one of the most frequently used tasks because it does not require extra annotations. This task aims to determine whether a sequence of frames from a video is in the correct temporal order. Inspired by that, we design two auxiliary tasks for temporal grounding models to promote the visual-textual matching features mining.
% Our method also applies this task to force the grounding model to learn the correct long-term temporal relations. Besides, we design a new auxiliary task to constrain the model to learn the true visual-textual matching features.

\section{Methods}

\subsection{Problem Formulation}
Given an untrimmed video $V$ and a sentence query $Q$, temporal grounding aims to determine the start and end timestamps $(\tau^s, \tau^e)$ of specific video moment semantically corresponding to the sentence query. The video $V$ is represented as $\boldsymbol{V}=\{\boldsymbol{v}_t\}^T_{t=1}$ frame-by-frame, and the query is represented as $\boldsymbol{Q}=\{\boldsymbol{q}_n\}^N_{n=1}$ word-by-word, where $T$ and $N$ are the number of frames and words, respectively.

\subsection{Input Construction}
\begin{figure}[!t]
\centering
\includegraphics[width=1.0\textwidth]{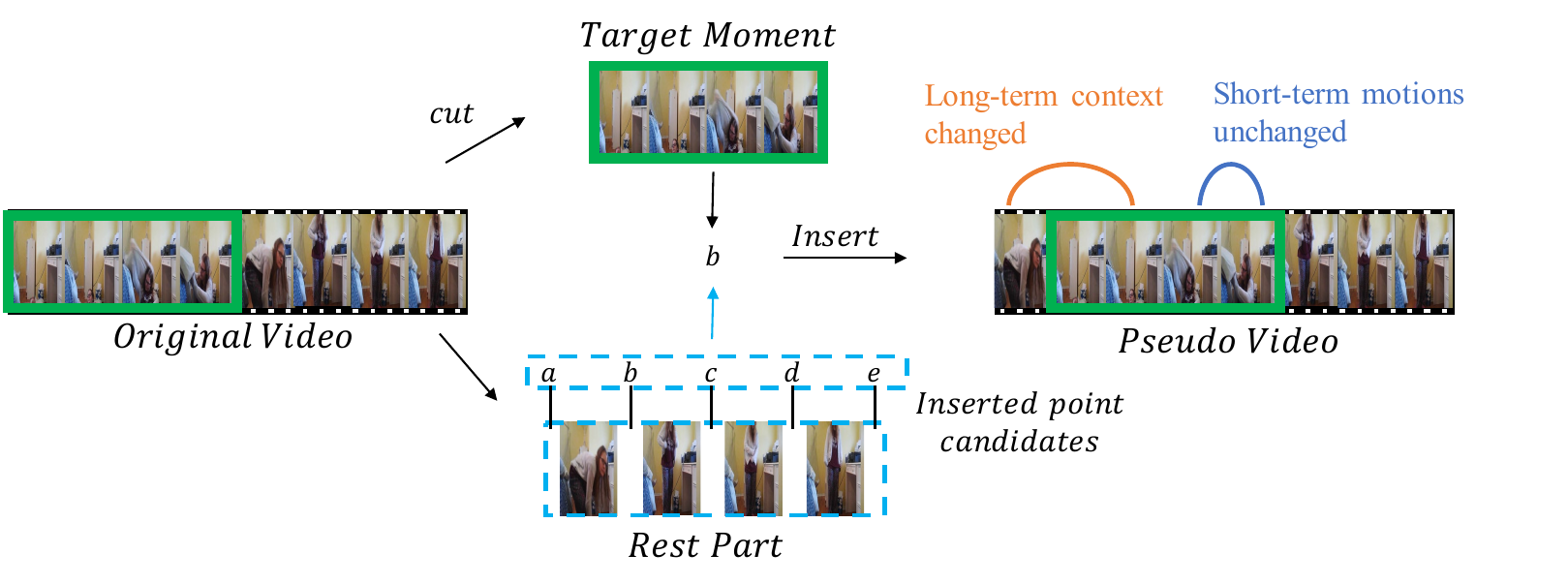}
\caption{An illustration of the generation of pseudo videos. The inserted point is randomly sampled from the candidates.}
\label{fig:pseudovideo}
\end{figure}

We first introduce how we shuffle videos to construct the input of our framework. For each video-query pair in the training set, we first cut out the target moment 
% corresponding to the query 
from the video and then insert the cut moment into a random temporal position of the rest of the video, as shown in Fig.~\ref{fig:pseudovideo}. We name the shuffled videos pseudo videos. The pseudo videos have three characteristics: 1) temporal positions of target moments do not match the temporal biases of queries; 2) spatial contents and short-term temporal motions within target moments are consistent with original videos; 3) long-term temporal contexts around target moments are disturbed. Leveraging the three characteristics of pseudo videos, we design two auxiliary tasks to suppress the effect of temporal biases on the final reasoning and strengthen the grounding model's perception ability on visual contents.

Formally, for each video-query pair $(V, Q)$ in the training set, we construct a triplet $(V, \bar{V}, Q)$ as the input. $\boldsymbol{\bar{V}}=\{\boldsymbol{\bar{v}}_t\}^T_{t=1}$ denotes the generated pseudo video and the corresponding timestamps of the target moment is $(\bar{\tau}^s, \bar{\tau}^e)$. Pseudo video $\bar{V}$ has the same length as original video $V$.

\begin{figure*}[!t]
\centering
\includegraphics[width=1.0\textwidth]{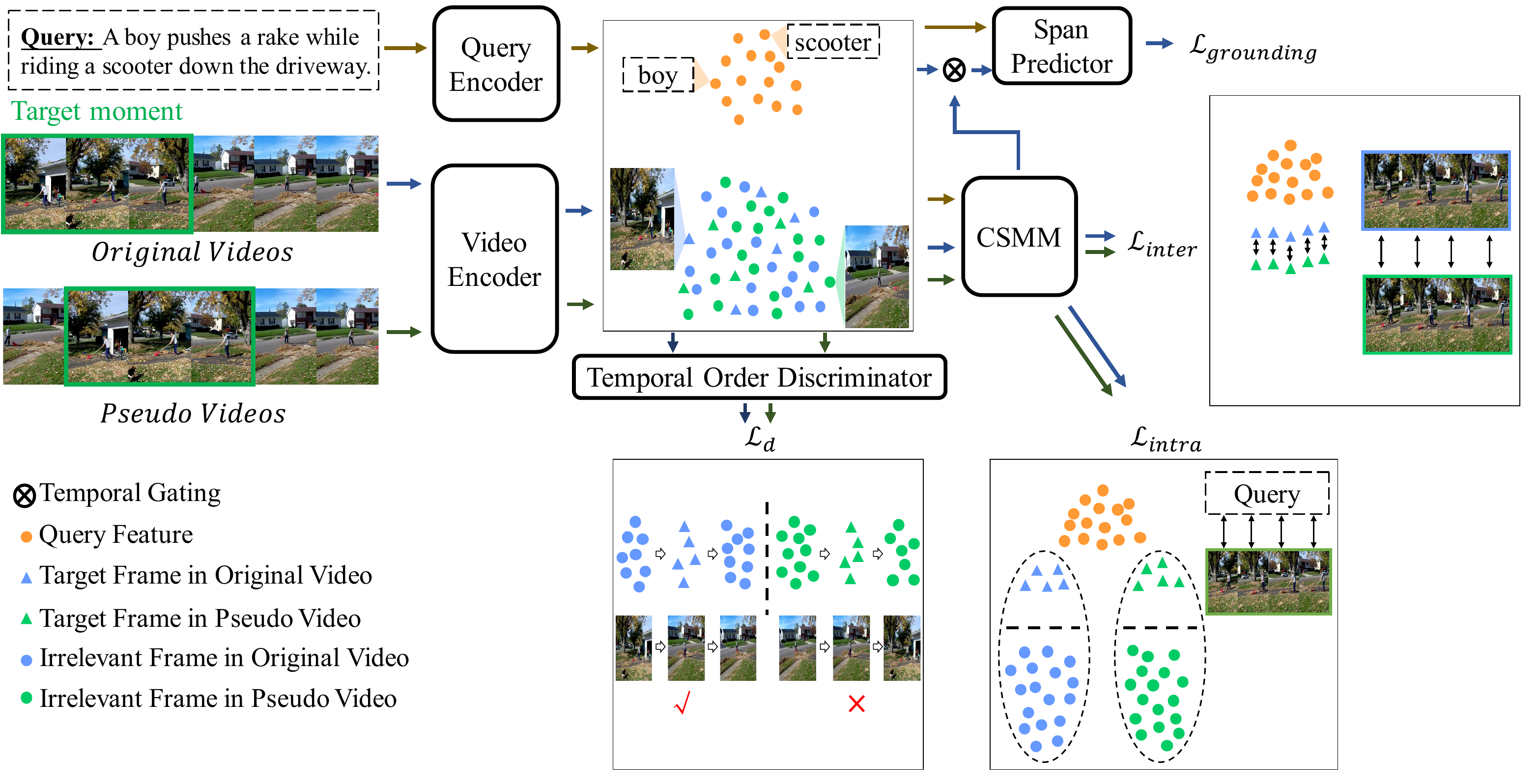}
\caption{An overview of our proposed framework with two auxiliary tasks. Video and query encoder encodes video and language modalities, respectively. The span predictor predicts the boundary scores for each frame. CSMM denotes the cross-modal semantic module, which predicts the relevance to the query for each video frame. We constrain the predict scores for both intra- and inter-video to deeply mine the visual-textual semantics relevance. Temporal order discriminator classifies whether the video moment sequence is in correct order.}
\label{fig:JSMD}
\end{figure*}

\subsection{Framework Architecture}
As shown in Fig.~\ref{fig:JSMD}, our model consists of a grounding model, a cross-modal semantic matching module, and a temporal order discriminator. The grounding model predicts the locations of target moments. The latter two modules aim to address two auxiliary tasks, cross-modal matching and temporal order discrimination, respectively. The cross-modal semantic matching module predicts the relevance to the query for each video frame. And the predicted frame-level relevance scores will be used to gate the encoded video features in the span predictor. The temporal order discriminator predicts whether the input video is in correct order. The three modules share a video encoder so that the auxiliary tasks can promote the grounding model training. 

\subsection{Cross-Modal Matching Task}\label{sec:csmm}
The cross-modal matching task aims to strengthen the visual-textual matching based on the content consistency between pseudo and original videos. We design two losses to constrain the predicted relevance scores for intra- and inter-video, respectively. For intra-video, the cross-modal semantic matching module should discriminate which frames are semantically related to the query for both pseudo and original videos. We implement this through two binary cross-entropy losses,

\begin{equation} \label{eq:loss_bce}
 \mathcal{L}_{\mathrm{BCE}}(\boldsymbol{V}) = - \sum_{\boldsymbol{v}_t}^{\boldsymbol{V}} p(\boldsymbol{v}_t)\log(c(\boldsymbol{v}_t))+(1-p(\boldsymbol{v}_t))\log(1-c(\boldsymbol{v}_t)),
\end{equation}

\begin{equation} \label{eq:loss_intra}
 \mathcal{L}_{intra} =  \frac{1}{2} ( \mathcal{L}_{\mathrm{BCE}}(\boldsymbol{V}) + \mathcal{L}_{\mathrm{BCE}}(\boldsymbol{\bar{V}})),
\end{equation}
where $p(\boldsymbol{v}_t)$ is set to 1 if frame $\boldsymbol{v}_t$ is within the target moments and 0 for otherwise, $c(\boldsymbol{v}_t)$ denotes the predicted cross-modal relevance scores for frame $\boldsymbol{v}_t$ with sigmoid activation. For the same query, target frames in pseudo and original videos may be at different temporal positions. Thus the model has to mine visual contents semantically matching the query. Since the long-term temporal context in videos may be incorrect, 
% (the long-term temporal contexts are disturbed in pseudo videos), 
this loss encourages the model to reason cross-modal relevance more based on spatial contents and short-term temporal motions. 
% the model has to concentrate more on spatial contents and short-term temporal motions to align with the query semantics.

For inter-video, we constrain that the predicted relevance distributions within target moments should be consistent between pseudo and original videos. We use a Kullback-Leibler divergence to constrain the fine-grained consistency,
%for inter videos, as follows,

\begin{equation} \label{eq:loss_local}
 \mathcal{L}_{inter} = D_{\mathrm{KL}} (\boldsymbol{c} \ || \ \boldsymbol{\bar{c}}),
\end{equation}
where $\boldsymbol{c}$ denotes the softmaxed relevance score vector of frames $\boldsymbol{v}_t$ from $\tau^s$ to $\tau^e$ timestep in video $\boldsymbol{V}$ and $\boldsymbol{\bar{c}}$ denotes the softmaxed relevance score vector of frames $\boldsymbol{\bar{v}}_t$ from $\bar{\tau}^s$ to $\bar{\tau}^e$ timestep in pseudo video $\boldsymbol{\bar{V}}$. 
% Note that the softmax function is performed within target moments. 
This inter-video loss constrains the relative relevance differences within target moments unchanged even though the external temporal context is changed. Thus this loss further emphasizes the impact of spatial and short-term temporal motion features.

\subsection{Temporal Order Discrimination Task}
The cross-modal matching task emphasizes the impact of spatial and short-term temporal motion features in final prediction, but we wish the grounding model is capable of understanding the long-term temporal contexts, which is important for temporal grounding task~\cite{mllc2018,2dtan2020,matn2021,smin2021}. However, some methods~\cite{mllc2018,2dtan2020,cbp2020,fvmr2021} learn the potential temporal position information during context capturing and thus suffer the temporal bias problem~\cite{bmvc2020,dcm2021}. To this end, we introduce a temporal order discrimination task to guide explicitly the learning of long-term temporal contexts. This task aims to discriminate whether the input video is in correct temporal order. Unlike the existing temporal order tasks~\cite{o3d,cliporder2019} that focus on the order of frames sampled from a short-span action, we design a task to focus on the long-term temporal context. Specifically, given a video-query pair, we divide the video into three parts: the target moment, the moment before the target moment, and the moment after the target moment, and ask whether the three moments are correctly ordered. We determine the supervision for this task based on the video type, i.e., we suppose that the orders of the original videos are correct and the ones of the pseudo videos are incorrect. This task is trained by a cross-entropy loss, which is denoted as $\mathcal{L}_{d}$.

\begin{equation} \label{eq:loss_d}
   L_{d} = - \sum_{c=1}^{C}{y_{V,c} \log{o_c(V)}} - \sum_{c=1}^{C}{y_{\bar{V},c} \log{o_c(\bar{V})}}
\end{equation}
where $C$ denotes the video categories (original or pseudo), $y$ is groundtruth label and $o_c(V)$ denotes the softmaxed prediction score of video $V$ for category $c$.

\subsection{Span-based Grounding Model}
We apply a span-based grounding model~\cite{qave} as our baseline model. It applies a typical span-based architecture, consisting of a query encoder, a video encoder, and a span predictor\footnote{As the grounding model is not our key contribution, more details about the grounding network and inference stage are provided in our supplementary material.}. Query encoder models the sentence query with multi-layered bidirectional LSTM with pre-trained language model (GloVe \cite{DBLP:conf/emnlp/PenningtonSM14}) embeddings as input. The encoded word-level embeddings and sentence-level representation are denoted as $\boldsymbol{W}=\{\boldsymbol{\dot{w}}_n\}^N_{n=1}$ and $\boldsymbol{s}$, respectively. The video encoder is guided by the query to encode video features over time. Different from \cite{qave}, we only use the word-level features $\boldsymbol{W}$ to guide the video encoding. 
\begin{equation} \label{eq:query_enc}
\boldsymbol{W}, \boldsymbol{s} = \mathrm{QueryEncoder} (\boldsymbol{Q}),  \quad
\boldsymbol{\dot{V}} = \mathrm{VideoEncoder} (\boldsymbol{V}, \boldsymbol{W}).
\end{equation}

Each encoded frame feature $\boldsymbol{\dot{v}}$ is concatenated with the sentence representation $\boldsymbol{s}$ and gated by the cross-modal relevance score $c(\boldsymbol{v}_t)$ before feed forward to the span predictor. The span predictor predicts the boundary scores $S_{start}(t)$, $S_{end}(t)$ for each frame. 

\begin{equation} \label{eq:start_predictor}
\begin{aligned}
 S_{start}(t) &= \mathrm{StartPredictor} (c(\boldsymbol{v}_t) (\boldsymbol{\dot{v}}_t || \boldsymbol{s})), \\
 S_{end}(t) &= \mathrm{EndPredictor} (c(\boldsymbol{v}_t) (\boldsymbol{\dot{v}}_t || \boldsymbol{s})).
\end{aligned}
\end{equation}

The start and end scores are normalized with SoftMax to obtain $P_{start}(t)$, $P_{end}(t)$, and trained using negative log-likelihood loss:
\begin{equation} \label{eq:loss_grounding}
 \mathcal{L}_g = -\log(P_{start}(t^s))-\log(P_{end}(t^e)),
\end{equation}
where $t^s$, $t^e$ are the ground-truth start and end frame indices for the original video, respectively. The ground-truth frame indices ($t^s$, $t^e$) are mapped from the ground-truth time values ($\tau^s, \tau^e$).

\subsection{Cross-Modal Semantic Matching Module}
The cross-modal semantic matching module predicts the relevance to the query for each video frame. The module is implemented by a multi-layered perceptron (MLP) with relu activation in hidden layers. We also use concatenation as the cross-model interaction method,

\begin{equation} \label{eq:cross_predictor}
 c(\boldsymbol{v}_t) = \boldsymbol{W}^c_2 \mathrm{relu}(\boldsymbol{W}^c_1 (\boldsymbol{\dot{v}}_t || \boldsymbol{s}) + \boldsymbol{b}^c_1) + \boldsymbol{b}^c_2,
\end{equation}
where $\boldsymbol{W}^c_1$, $\boldsymbol{W}^c_2$, $\boldsymbol{b}^c_1$ and $\boldsymbol{b}^c_2$ are the learnable parameters of MLP and shared across all time steps.
We also apply a temporal gating on the encoded video features using the predicted relevance scores to highlight the impact of the matching results on the final reasoning, as shown in Eqs.~(\ref{eq:start_predictor}).

\subsection{Temporal Order Discriminator}
Given a moment tuple $(M_1=\{\boldsymbol{\dot{v}}_1, \cdots, \boldsymbol{\dot{v}}_{\tau^s-1}\}, M_2=\{\boldsymbol{\dot{v}}_{\tau^s}, \cdots, \boldsymbol{\dot{v}}_{\tau^e}\}, M_2=\{\boldsymbol{\dot{v}}_{\tau^e+1}, \cdots, \boldsymbol{\dot{v}}_{T}\})$, we first obtain the moment-level represenstations for these moments by average pooling the encoded frame features within the moments,

\begin{equation} \label{eq:momentfeat}
   \boldsymbol{m}_{1} =  \mathrm{pooling}(M_1), \ \boldsymbol{m}_{2} =  \mathrm{pooling}(M_2), \ \boldsymbol{m}_{3} =  \mathrm{pooling}(M_3).
\end{equation}
We mainly focus on the contexts around the target moment~($M_2$) to reason the order correctness. We concatenate the paired moment representations and use two parallel fully-connected layers with shared parameters to obtain the context information $\boldsymbol{h}_1$, $\boldsymbol{h}_2$, respectively,

\begin{equation} \label{eq:fb}
 \boldsymbol{h}_{1} =  \mathrm{relu}(\boldsymbol{W}^o_1 (\boldsymbol{m}_1 || \boldsymbol{m}_2 + \boldsymbol{b}^o_1)), \ 
 \boldsymbol{h}_{2} =  \mathrm{relu}(\boldsymbol{W}^o_1 (\boldsymbol{m}_2 || \boldsymbol{m}_3 + \boldsymbol{b}^o_1)).
\end{equation}
Then we concatenate the context information with the target moment representation and predict the classification scores,
% that the moment tuple belongs to the original videos or the shuffled videos,

\begin{equation} \label{eq:tod}
 o(\boldsymbol{V}) = \boldsymbol{W}^o_2 (\boldsymbol{m}_2 || \boldsymbol{h}_{1} || \boldsymbol{h}_{2}) + \boldsymbol{b}^o_2,
\end{equation}
where $\boldsymbol{W}^o_1$, $\boldsymbol{W}^o_2$, $\boldsymbol{b}^o_1$ and $\boldsymbol{b}^o_2$ are the learnable parameters of fully-connected layers and shared across videos. To prevent the learning of temporal biases, we reason the order correctness only based on the content relevance and do not introduce any global temporal position information. The position information contained in the encoded frame features is diluted by the average-pooling operation.

\subsection{Training Objective}
The final training loss is the weighted summarization of the loss of each module,
\begin{equation} \label{eq:loss}
 \mathcal{L} = \mathcal{L}_{g} + \lambda_1 \mathcal{L}_{intra} + \lambda_2 \mathcal{L}_{inter} + \lambda_3 \mathcal{L}_{d}.
\end{equation}

\section{Experiments}
\subsection{Datasets}

\paragraph{Charades-STA} Charades-STA is built on Charades dataset~\cite{DBLP:conf/eccv/SigurdssonVWFLG16} by \cite{tall2017}. The videos in this dataset are mainly about indoor activities. The average length of videos and annotated moments are 30s and 8s, respectively.

\paragraph{ActivityNet Captions} ActivityNet Captions is built on ActivityNet~\cite{caba2015activitynet}, which is a large-scale dataset of human activities based on YouTube videos. The average length of videos and the annotated moments are 117s and 36s, respectively.

\subsection{Dataset Splits}
In the original splits of the two datasets, the training and test sets have similar temporal biases. Thus the methods that learn the temporal biases in the training set could also perform well on the test set~\cite{bmvc2020,closelook2020}. To eliminate the impact of temporal bias problem on evaluation performance, we perform experiments on the re-divided splits\footnote{\url{https://github.com/yytzsy/grounding_changing_distribution}} proposed by \cite{closelook2020}. In the re-divided splits, each dataset is re-divided into four sets: training, validation(val), test-iid, and test-ood. The temporal locations of all samples in the training, val, and test-iid satisfy the independent and identical distribution, and the samples in test-ood are out-of-distribution. Therefore, it is useless to exploit the temporal biases in training set to make predictions on test-ood set. The test-ood sets on both datasets have similar vocabulary distributions to the training set, which means that the difference of the temporal distribution between the training and test-ood sets is the main challenge of the re-divided splits. The sample statistics of the two splits are reported in Table~\ref{tab:datasets}.

% \usepackage{multirow}
\begin{table}[t]
\scriptsize
\centering
\caption{The statistics of the number of videos and query-moment pairs in different datasets and splits.}
\label{tab:datasets}
\begin{tabular}{|c|ccc|ccc|}
\hline
\multirow{2}{*}{Dataset} & \multicolumn{3}{c|}{Original Splits}                              & \multicolumn{3}{c|}{Re-divided Splits}                               \\ \cline{2-7} 
                         & \multicolumn{1}{c|}{Split} & \multicolumn{1}{c|}{Videos} & Pairs  & \multicolumn{1}{c|}{Split}    & \multicolumn{1}{c|}{Videos} & Pairs  \\ \hline
\multirow{4}{*}{Charades-STA} &
  \multicolumn{1}{c|}{training} &
  \multicolumn{1}{c|}{5,338} &
  12,408 &
  \multicolumn{1}{c|}{training} &
  \multicolumn{1}{c|}{4,564} &
  11,071 \\
                         & \multicolumn{1}{c|}{test}  & \multicolumn{1}{c|}{1,334}  & 3,720  & \multicolumn{1}{c|}{val}      & \multicolumn{1}{c|}{333}    & 859    \\
                         & \multicolumn{1}{c|}{}      & \multicolumn{1}{c|}{}       &        & \multicolumn{1}{c|}{test-iid} & \multicolumn{1}{c|}{333}    & 823    \\
                         & \multicolumn{1}{c|}{}      & \multicolumn{1}{c|}{}       &        & \multicolumn{1}{c|}{test-ood} & \multicolumn{1}{c|}{1,442}  & 3,375  \\ \hline
\multirow{4}{*}{ActivityNet Captions} &
  \multicolumn{1}{c|}{training} &
  \multicolumn{1}{c|}{10,009} &
  37,421 &
  \multicolumn{1}{c|}{training} &
  \multicolumn{1}{c|}{10,984} &
  51,415 \\
                         & \multicolumn{1}{c|}{val}   & \multicolumn{1}{c|}{4,917}  & 17,505 & \multicolumn{1}{c|}{val}      & \multicolumn{1}{c|}{746}    & 3,521  \\
                         & \multicolumn{1}{c|}{test}  & \multicolumn{1}{c|}{4,885}  & 17,031 & \multicolumn{1}{c|}{test-iid} & \multicolumn{1}{c|}{746}    & 3,443  \\
                         & \multicolumn{1}{c|}{}      & \multicolumn{1}{c|}{}       &        & \multicolumn{1}{c|}{test-ood} & \multicolumn{1}{c|}{2,450}  & 13,578 \\ \hline
\end{tabular}
\end{table}

\subsection{Experimental Settings}
\paragraph{Metrics} Following the conventions, we adopt R@$n$, IoU=$\theta$ and mIoU as evaluation metrics. R@$n$, IoU=$\theta$ represents the percentage of testing samples having at least one result whose IoU with ground truth is larger than $\theta$ in top-$n$ localized results. mIoU represents the average IoU over all testing samples. Following previous works~\cite{excl2019,cbp2020,vslnet2020}, we use $n = 1$ and $\theta \in \{0.3, 0.5, 0.7\}$.

\paragraph{Implementation}
For natural language, we use 300d Glove \cite{DBLP:conf/emnlp/PenningtonSM14} vectors as word embeddings. 
For the words not in the vocabulary of Glove, we generate their embeddings randomly. 
% The vocabulary sizes are 1,294 and 13,745 for Charades-STA and ActivityNet Captions datasets, respectively. We truncate all sentence queries with a maximum of 25 words for ActivityNet Captions dataset and 15 for Charades-STA dataset.
For video modality, we use 1024d I3D feature pre-trained on Kinetics dataset \cite{kinetics} or 500d C3D feature \cite{c3d} pre-trained on Sports-1M dataset \cite{sports1m} as the initial frame features, and downsample the videos at a frame rate of 1 frame per second.
% We set the length of video feature sequences to 128 for Charades-STA and 240 for ActivityNet Captions, respectively. Longer is truncated, and shorter is padded. 
For each training epoch, we will re-generate a pseudo video for each video-query pair. We train the model with a batch size of 32 for 30 epochs for all datasets using Adam optimizer with an initial learning rate of 0.001. We set $\lambda_1$, $\lambda_2$ and $\lambda_3$ in Eq.~\ref{eq:loss} to 1, 1, 1, respectively. More details are provided in supplementary material.

% \usepackage{booktabs}
% \usepackage{multirow}
% \usepackage[normalem]{ulem}
\begin{table}[t]
\scriptsize
\caption{Comparison on Charades-CD split using I3D features.}
\label{tab:charades_cd}
\centering
\begin{tabular}{|c|cccc|cccc|}
\hline% \toprule
\multirow{2}{*}{Model}   & \multicolumn{4}{c|}{test-iid}       & \multicolumn{4}{c|}{test-ood}        \\ \cline{2-9}%\cmidrule(l){2-9} 
                         & IoU=0.3 & IoU=0.5 & IoU=0.7 & mIoU  & IoU=0.3 & IoU=0.5 & IoU=0.7 & mIoU  \\ \hline%\midrule
2D-TAN~\cite{2dtan2020}  & 60.15   & 49.09   & 26.85   & 42.73 & 52.79   & 35.88   & 13.91   & 34.22 \\
LG~\cite{lg2020}         & 64.52   & 51.28   & 28.68   & 45.16 & 59.32   & 42.90   & 19.29   & 39.43 \\
DRN~\cite{drn2020}       & 53.22   & 42.04   & 23.32   & 28.21 & 45.87   & 31.11   & 15.17   & 23.05 \\
VSLNet~\cite{vslnet2020} & 61.48   & 43.26   & 28.43   & 42.92 & 54.61   & 34.10   & 17.87   & 36.34 \\
DCM~\cite{dcm2021} & {\ul 67.27}    & {\ul 55.81}    & {\ul 37.30}    & {\ul 48.74}    & {\ul 60.89}    & {\ul 45.47}    & {\ul 22.70}    & {\ul 40.99}    \\
Baseline                 & 66.34   & 50.55   & 34.26   & 46.81 & 58.96   & 38.22   & 20.50   & 39.52 \\
\textbf{Ours}      & \textbf{70.72} & \textbf{57.59} & \textbf{37.79} & \textbf{50.93} & \textbf{64.95} & \textbf{46.67} & \textbf{27.08} & \textbf{44.30} \\ 
\hline % \bottomrule
\end{tabular}
\end{table}

\subsection{Comparison with State-of-the-Arts}
We compare our methods with the recently state-of-the-art methods and our baseline, which only contains a span-based grounding model~\cite{qave} and is only supervised by the grounding loss $\mathcal{L}_{g}$. We first analyze the competitors' performance on the re-divided splits. Then we follow \cite{bmvc2020} and perform a test to check whether the competitors suffer the temporal bias problem. We also show the performance comparison on the original splits.

% \usepackage{booktabs}
% \usepackage{multirow}
% \usepackage[normalem]{ulem}
\begin{table}[t]
\scriptsize
\caption{Comparison on ActivityNet-CD split using I3D features.}
\label{tab:anet_cd}
\centering
\begin{tabular}{|c|cccc|cccc|}
% \toprule
\hline
\multirow{2}{*}{Model}   & \multicolumn{4}{c|}{test-iid}                         & \multicolumn{4}{c|}{test-ood}                          \\ \cline{2-9}%\cmidrule(l){2-9} 
                         & IoU=0.3     & IoU=0.5     & IoU=0.7     & mIoU        & IoU=0.3     & IoU=0.5     & IoU=0.7     & mIoU        \\ \hline %\midrule
2D-TAN~\cite{2dtan2020}  & 60.56       & 46.59       & 30.55       & 44.99       & 40.13       & 22.01       & 10.34       & 28.31       \\
LG~\cite{lg2020}         & 61.63       & 46.41       & 29.28       & 44.62       & {\ul 40.78} & {\ul 23.85} & 10.96       & {\ul 28.46} \\
VSLNet~\cite{vslnet2020} & {\ul 62.71} & {\ul 47.81} & 29.07       & {\ul 46.33} & 38.30       & 20.03       & 10.29       & 28.18       \\
DCM~\cite{dcm2021}       & 60.15       & 47.26       & {\ul 31.97} & 45.20       & 39.39       & 22.32       & {\ul 11.22} & 28.08       \\
Baseline                 & 60.56       & 44.58       & 27.42       & 44.28       & 38.78       & 21.39       & 10.86       & 28.41       \\
\textbf{Ours} & \textbf{63.29} & \textbf{48.07} & \textbf{32.15} & \textbf{47.03} & \textbf{42.08} & \textbf{24.57} & \textbf{13.21} & \textbf{30.45} \\
\hline
% \bottomrule
\end{tabular}
\end{table}

\subsubsection{Comparison on the Re-divided Splits.}
Table~\ref{tab:charades_cd} and Table~\ref{tab:anet_cd} summarize the results on the splits Charades-CD and ActivityNet-CD, respectively. For a fair comparison, all methods use the same pre-trained visual features. Best results are in \textbf{bold} and second-best \underline{underlined}. Observed that all methods have a significant performance drop on the test-ood compared to the test-iid on both datasets. The change of the temporal distribution on the test-ood set challenges the model's generalization ability. Compared with the state-of-the-art methods, our method performs best over all evaluation metrics on both test sets and on both datasets. Particularly, our method outperforms all methods by clear margins on the test-ood splits, especially in the metrics IoU=0.7 and mIoU. It shows that our method has a stronger generalization ability against the temporal distributions. Besides, observed that our method improves the baseline on both test sets, but we achieve more relative improvements on test-ood set than test-iid set, e.g., 22.11\% v.s. 13.93\% in IoU=0.5 on Charades-CD and 14.87\% v.s. 7.83\% on ActivityNet-CD. It shows the effectiveness of our method on strengthening the model's generalization ability against the different temporal distributions. 

\begin{figure}[t]
\centering
 \begin{subfloat}[R@1~(IoU$=$0.5) scores for competitors when the raw input videos and randomized ones are fed into these models.]{\includegraphics[width=0.48\textwidth]{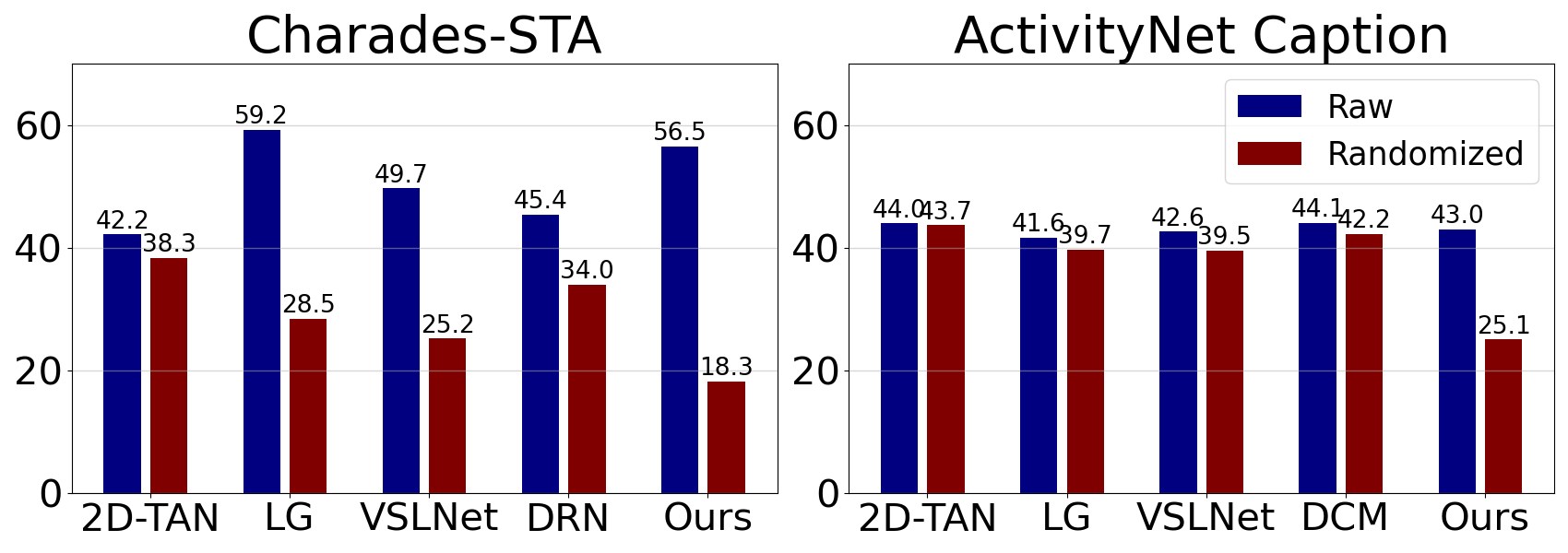}%
  \label{fig:visualinput}}
 \end{subfloat}
 \begin{subfloat}[Comparison with the state-of-the-arts on the original splits of Charades-STA and ActivityNet Captions.]{\includegraphics[width=0.48\textwidth]{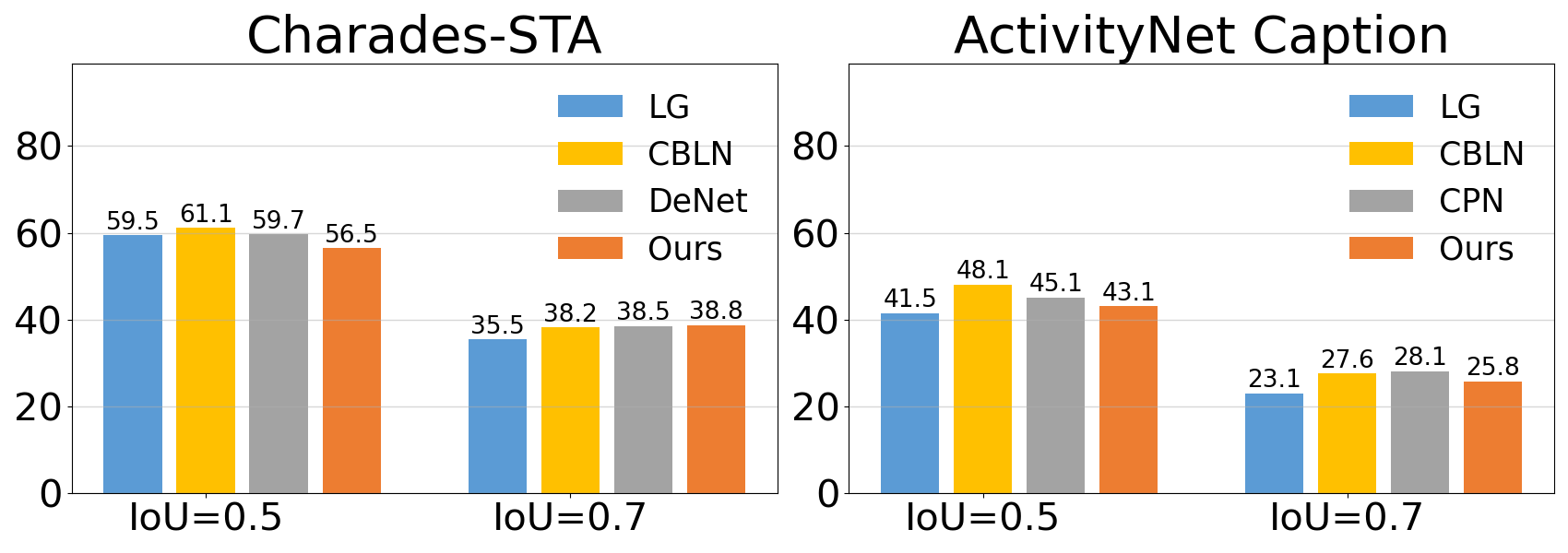}%
  \label{fig:sota}}
 \end{subfloat}
\caption{Sanity check on visual input~(a) and comparison on the original splits~(b)}
\label{fig:visualandsota} 
\end{figure}

\subsubsection{Sanity Check on Visual Input.}
 Suffering temporal bias problem, a model can ignore the visual input but perform well on the evaluation metrics on the original splits. Thus, same as~\cite{bmvc2020}, we perform a test on some state-of-the-art competitors\footnote{We used the models with trained parameters provided by their authors if available.} to show how much these models take input videos into account for prediction. Specifically, we divide input videos into short segments and randomly reorder them before evaluating the models. This randomization messes up the correspondence between input videos and ground truth temporal locations. If a model makes the prediction based on visual input, the performance should drop significantly by the randomization; otherwise, we can conclude that the model doesn't make use of the visual input during reasoning. 
 % Further, if the prediction results are similar to the raw video, we can conclude that the model makes predictions based on the temporal biases of queries.
 Fig.~\ref{fig:visualinput} shows the results for the competitors and our method. On Charades-STA, all state-of-the-arts except 2D-TAN~\cite{2dtan2020} and DRN~\cite{drn2020} show significant performance drops for randomized videos. Among all competitors, our method performs the most significant drop, which demonstrates the effectiveness of our method for addressing temporal bias problem on this dataset. On ActivityNet Captions, all state-of-the-arts including DCM~\cite{dcm2021} achieve similar performance using randomized videos to the raw ones, which shows that these methods actually do not use the visual input and over-rely on the temporal biases. Our method performs a clear drop in this test on ActivityNet Captions, which validates that our method can make the model focus more on the visual content and mitigate the reliance on the temporal biases on this dataset.

\subsubsection{Comparison on the Original Splits.} We also compare our method with the state-of-the-arts~\cite{lg2020,cbln2021,denet2021,cpn2021} on the original splits of the two datasets. For a fair comparison, our method uses i3d features on Charades-STA and c3d on ActivityNet Captions. As shown in Fig.~\ref{fig:sota}, our method achieves competitive performance to the state-of-the-arts on both datasets, which shows that our method can address the temporal bias problem without loss in grounding accuracy.

\subsection{Ablation Study}

% \usepackage{multirow}
\begin{table}[t]
\scriptsize
\caption{Comparison of different usage of pseudo videos on test-ood sets.}
\label{tab:pseudovideo}
\begin{tabular}{|c|cccc|cccc|}
\hline
\multirow{2}{*}{Methods} & \multicolumn{4}{c|}{Charades-CD}    & \multicolumn{4}{c|}{ActivityNet-CD} \\ \cline{2-9} 
                         & IoU=0.3 & IoU=0.5 & IoU=0.7 & mIoU  & IoU=0.3 & IoU=0.5 & IoU=0.7 & mIoU  \\ \hline
Baseline                 & 58.96   & 38.22   & 20.50   & 39.52 & 38.78   & 21.39   & 10.86   & 28.41 \\
Baseline+DataAug         & 58.61   & 37.40   & 19.82   & 39.06 & 40.74   & 22.27   & 12.16   & 29.59 \\
Baseline+CSMM+DataAug    & 59.22   & 38.45   & 21.41   & 39.84 & 28.42   & 13.36   & 6.13    & 20.29 \\
Ours                     & 64.95   & 46.67   & 27.08   & 44.30 & 42.08   & 24.57   & 13.21   & 30.45 \\ \hline
\end{tabular}
\end{table}

\subsubsection{Pseudo videos.}
We study the effect of the pseudo videos. We first test applying the data augmentation strategy based on the baseline, i.e., treating the pseudo and original videos equally as training samples. Then we add the cross-modal matching module and use the frame-level relevance scores to temporally gating the encoded video features. The relevance scores are supervised by Eq.(\ref{eq:loss_bce}). As shown in Table~\ref{tab:pseudovideo}, after applying data augmentation, there are slight improvements from the baseline on ActivityNet-CD while the performance is inferior to the baseline on Charades-CD. After adding CSMM, there are slight improvements on Charades-CD but a significant performance drop on ActivityNet-CD. On the contrary, our method improves the performance from the baseline with clear margins on both datasets. The comparisons validate the infeasibility of shuffled videos as augmented training samples and the superiority of our method.

\begin{table}[t]
\centering
\scriptsize
\caption{Ablation study of loss terms on test-ood sets.}
\label{tab:loss_terms}
\begin{tabular}{|c|cccc|cccc|cccc|}
\hline%\toprule
\multirow{2}{*}{Row} &
  \multicolumn{4}{c|}{Loss Terms} &
  \multicolumn{4}{c|}{Charades-CD} &
  \multicolumn{4}{c|}{ActivityNet-CD} \\ \cline{2-13}%\cmidrule(l){2-13} 
 &
  $\mathcal{L}_{g}$ &
  $\mathcal{L}_{intra}$ &
  $\mathcal{L}_{inter}$ &
  $\mathcal{L}_{d}$ &
  IoU=0.3 &
  IoU=0.5 &
  IoU=0.7 &
  mIoU &
  IoU=0.3 &
  IoU=0.5 &
  IoU=0.7 &
  mIoU \\ \hline%\midrule
1 & \Checkmark &            &            &            & 55.03 & 28.91 & 14.08 & 35.05 & 37.30 & 20.30 & 10.42 & 26.75 \\
2 & \Checkmark & \Checkmark &            &            & 62.34 & 44.60 & 25.41 & 42.66 & 40.39 & 22.45 & 11.51 & 29.77 \\
3 & \Checkmark &            & \Checkmark &            & 55.39 & 33.13 & 17.23 & 36.36 & 34.85 & 15.80 & 7.66  & 26.27 \\
4 & \Checkmark &            &            & \Checkmark & 55.91 & 6.16  & 2.79  & 30.47 & 34.84 & 15.65 & 7.95  & 26.43 \\
5 & \Checkmark & \Checkmark &            & \Checkmark & 62.96 & 45.36 & 26.73 & 43.40 & 41.49 & 23.12 & 12.55 & 30.04 \\
6 & \Checkmark & \Checkmark & \Checkmark &            & 62.79 & 44.23 & 25.90 & 42.93 & 41.33 & 23.28 & 12.53 & 30.04 \\
7 & \Checkmark & \Checkmark & \Checkmark & \Checkmark & 64.95 & 46.67 & 27.08 & 44.30 & 42.08 & 24.57 & 13.21 & 30.45 \\ \hline%\bottomrule
\end{tabular}
\end{table}

\subsubsection{Loss terms.}
We analyze the impact of each loss term and their combinations. Table~\ref{tab:loss_terms} summarizes the results, and some reveal points are listed as follows.

(1) $\mathcal{L}_{intra}$ leads to the main performance boosts on test-ood sets (comparing Rows 1, 2 and 7). It validates our design of using the content consistency between shuffled and original videos to mitigate the reliance on temporal biases.

(2) Without $\mathcal{L}_{intra}$, the improvement of adding $\mathcal{L}_{inter}$ is limited~(comparing Rows 1 and 3). It means that only constraining the relative relevance differences within target moments cannot highlight the target moment from the entire video. The combination of $\mathcal{L}_{intra}$ and $\mathcal{L}_{inter}$ can further improve the performance (comparing Rows 2 and 6), which validates the effectiveness of our design of contraining the cross-modal relevance scores between shuffled and original videos.

(3) Only adding $\mathcal{L}_{d}$ leads to performance drops on both datasets, especially on the high IoU  (comparing Rows 1 and 4).
The baseline over-relies on memorizing temporal biases but the temporal order discrimination task restricts the learning of temporal position information. So the baseline model degrades to predict long-span predictions\footnote{The length distribution of predictions can be found in our supplementary material.}. After adding the cross-modal matching task, the supervision guides the model to focus on the short-term visual contents semantically matching queries and thus leads to performance boosts (Rows 5 and 7).

\begin{figure*}[!t]
\centering
\includegraphics[width=0.65\textwidth]{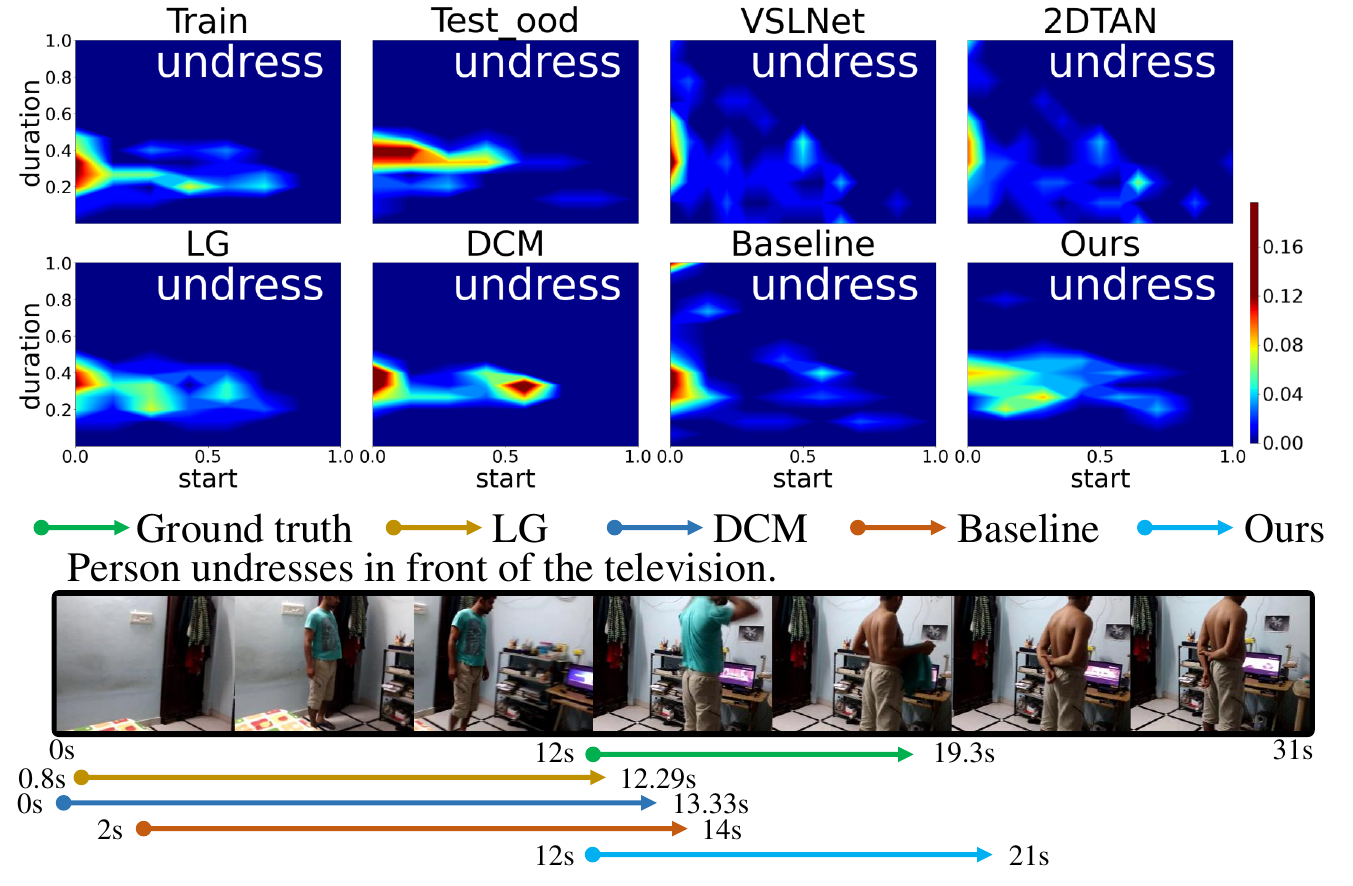}
\caption{Top is the temporal distribution comparison of the word `undress' on Charades-CD. Color represents value of probability. Bottom is an example of grounding results for a query containing the word 'underss'.}
\label{fig:visual}
\end{figure*}

\subsubsection{Visualization.}
We show a qualitative example on Charades-CD to show how models suffer the temporal bias problem and the effect of our method in Fig.~\ref{fig:visual}. We first show the temporal distribution comparison of the word `undress' between the training/test\_ood sets and the prediction results of grounding models on test\_ood set. 
% On the training set, target moments are biased to start at the beginning of videos and span approximately one-fifth to one-half of video duration. 
Observed that the competitor models have significant and similar biases to the training set while our model does not. Then we show a test samples of the word `undress' and the grounding results of different methods. In this sample video, groundtruth target moment does not start at the beginning. 
% (the person begins to undress in 12s).
But most models (LG~\cite{lg2020}, DCM~\cite{dcm2021}, and our baseline) still make the predictions fitting the biases in training set. With the training of our framework, we can effectively 
mitigate the baseline's reliance on biases and 
turn the model's attention back to visual contents to make correct predictions. We provide more qualitative examples in our supplementary material.

% We show several qualitative grounding examples to show how models suffer the temporal bias problem and the effect of our method in Fig.~\ref{fig:visual}. We first show three words' distribution maps in the training set of Charades-CD. Observed that target moments corresponding to the words `undress', `hold', and `awaken' are biased to start at the beginning of videos and span approximately one-fifth to one-half of video duration. Then we show the test samples whose queries contain one of the three words and the corresponding grounding results produced by different methods. In these sample videos, groundtruth target moments do not start at the beginning (e.g., in example (a), the person begins to undress in 12s of the video). But most methods (LG~\cite{lg2020}, DCM~\cite{dcm2021}, and our baseline) ignore the visual content and still make the predictions fitting the biases in training set as shown in Fig.~\ref{fig:visual}. Observed that, with the training of our framework, we can effectively mitigate the baseline's reliance on biases and turn the model's attention back to visual contents to make correct predictions.

\section{Conclusion}
This paper proposes a novel training framework for temporal grounding models to leverage shuffled videos to address the temporal bias problem. We propose two auxiliary tasks to suppress the effect of temporal biases and strengthen the model's perception ability on visual contents. Extensive experiments on Charades-STA and ActivityNet Captions demonstrate the effectiveness of our method on strengthening the generalization ability of the grounding model and mitigating the reliance on temporal biases.

\subsection*{Acknowledgement}
This work was supported in part by the National Natural Science Foundation of China under Grants (62071067, 62001054, 62101064, 62171057), in part by the Ministry of Education and China Mobile Joint Fund (MCM20200202), China Postdoctoral Science Foundation under Grant 2022M710468, Beijing University of Posts and Telecommunications-China Mobile Research Institute Joint Innovation Center.

\clearpage
\bibliographystyle{splncs04}
\bibliography{CanShufflingVideo}
\end{document}